\pdfoutput=1

\documentclass[11pt]{article}

\usepackage[preprint]{acl}

\usepackage{times}
\usepackage{latexsym}

\usepackage[T1]{fontenc}
\usepackage{CJKutf8}

\usepackage[utf8]{inputenc}

\usepackage{microtype}

\usepackage{inconsolata}

\usepackage{graphicx}

\usepackage{tabularx}
\usepackage{booktabs}
\usepackage{multirow}
\usepackage{makecell}
\usepackage{adjustbox}
\usepackage{tikz}

\usepackage{subfig}
\usepackage{graphicx}
\usepackage{dsfont}
\usepackage{amsfonts}
\usepackage{amsmath}
\usepackage{stackrel}
\DeclareMathOperator*{\argmax}{arg\,max}

\usepackage[english]{babel}

\newcommand{\vx}{\mathbf{x}}
\newcommand{\vy}{\mathbf{y}}
\newcommand{\vh}{\mathbf{h}}

\newcommand{\dmbr}{MBR-OT}

\newcommand{\wdmbr}{MBR-WD}
\newcommand{\wdmbrl}{MBR-WD$_L$}
\newcommand{\ewdmbr}{MBR-WD$^\epsilon$}
\newcommand{\ewdmbrl}{MBR-WD$_L^\epsilon$}

\newcommand{\Pmodel}{P_\mathrm{{model}}} 
\newcommand{\Pm}{P}
\newcommand{\CandH}{\mathcal{H}_{\mathrm{cand}}}
\newcommand{\RefH}{\mathcal{H}_{\mathrm{ref}}}
\newcommand{\jp}[1]{\begin{CJK}{UTF8}{ipxm}#1\end{CJK}}

\usepackage{xcolor}

\title{Document-Level Text Generation with\\
Minimum Bayes Risk Decoding using Optimal Transport}

\author{Yuu Jinnai \\
  CyberAgent \\
  \texttt{jinnai\_yu@cyberagent.co.jp}}

\begin{document}
\maketitle
\begin{abstract}
Document-level text generation tasks are known to be more difficult than sentence-level text generation tasks as they require the understanding of longer context to generate high-quality texts.
In this paper, we investigate the adaption of Minimum Bayes Risk (MBR) decoding for document-level text generation tasks. 
MBR decoding makes use of a utility function to estimate the output with the highest expected utility from a set of candidate outputs.
Although MBR decoding is shown to be effective in a wide range of sentence-level text generation tasks, its performance on document-level text generation tasks is limited as many of the utility functions are designed for evaluating the utility of sentences.
To this end, we propose \dmbr{}, a variant of MBR decoding using Wasserstein distance to compute the utility of a document using a sentence-level utility function.
The experimental result shows that the performance of \dmbr{} outperforms that of the standard MBR in document-level machine translation, text simplification, and dense image captioning tasks. Our code is available at \url{https://github.com/jinnaiyuu/mbr-optimal-transport}.
\end{abstract}

\section{Introduction}

Large language models (LLMs) have demonstrated remarkable capabilities across various natural language processing tasks \cite{stiennon2020,NEURIPS2022_b1efde53,touvron2023llama,dubey2024llama,openai2024gpt4}. While many text-generation tasks are evaluated at the sentence level, LLMs are also capable of generating text at the document level \cite{wang-etal-2023-document-level,xia2024docgenome,zhang-etal-2024-bench}. This raises the need for evaluating the performance of decoding algorithms for document-level text generation tasks.

Minimum Bayes Risk (MBR) decoding has been shown to be highly effective for sentence-level directed text generation tasks \cite{goel2000minimum,freitag-etal-2022-high,eikema-aziz-2022-sampling,freitag-etal-2023-epsilon}. However, its effectiveness for generating longer texts is less investigated. In this paper, we evaluate the performance of MBR decoding in document-level text generation tasks. 
Specifically, we propose \dmbr{}, a variant of MBR decoding that uses an optimal transport distance as a utility function.
Optimal transport has been used in many fields to measure the dissimilarity of two probability distributions \cite{peyre2020computational,villani2021topics}. 
We use optimal transport as a tool to evaluate the utility between documents using sentence-level utility functions.
This approach enables the use of the sentence-level utility functions that have been investigated for years to improve their accuracy by many researchers \cite{freitag-etal-2022-results,freitag-etal-2023-results,freitag-etal-2024-llms}.

We evaluate \dmbr{} on multiple directed text generation tasks: document-level machine translation, document-level text simplification, and dense image captioning. Our results show that MBR decoding consistently outperforms the baselines across these tasks, showing its effectiveness for document-level text generation tasks. 

\begin{figure}
    \centering
    \adjustbox{max width=0.96\columnwidth}{
    \begin{tikzpicture}

    \node (text1) at (0,1) {I love cats. I love dogs.};
    \node (text2) at (0,-1) {I love dogs. I love cats.};

    \node (A1) at (-1.0,0.4) {A};
    \node (B1) at (1.0,0.4) {B};
    \node (B2) at (-1.0,-1.6) {B};
    \node (A2) at (1.0,-1.6) {A};

    \draw[->] (A1) -- (-1.0,0.8);
    \draw[->] (B1) -- (1.0,0.8);
    \draw[->] (B2) -- (-1.0,-1.2);
    \draw[->] (A2) -- (1.0,-1.2);
    
    \node[draw, circle] (A) at (3,1) {A};
    \node[draw, circle] (B) at (5,1) {B};

    \node[draw, circle] (C) at (3,-1) {B};
    \node[draw, circle] (D) at (5,-1) {A};

    \draw[dotted] (A) -- (D)  node[midway, above] {u=1};
    \draw[dotted] (B) -- (C)  node[midway, below] {u=1};

    \node at (5.3,0) {\textbf{WD = 0}};

\end{tikzpicture}
    }
    \caption{Illustrative example of a metric using Wasserstein distance over two texts "I love cats. I love dogs" and "I love dogs. I love cats.". Each output is segmented into a set of segments (e.g., sentence) and a utility function is used to compute the utility over a pair of segments from each of the outputs. Wasserstein distance is flexible to the change in the structure of the text, making it an adaptive measure for a wide range of tasks.}
    \label{fig:illustration}
\end{figure}
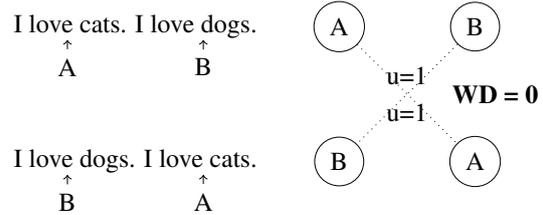

\section{Background}

We introduce MBR decoding as one of the algorithms to solve the task. Then, we explain the concept of optimal transport which is used by the proposed method.



\subsection{Minimum Bayes Risk (MBR) Decoding}

Unlike maximum a posteriori (MAP) decoding \cite{eikema-aziz-2020-map,Holtzman2020The}, which aims to find the most probable output, minimum Bayes risk (MBR) decoding selects the output that maximizes the expected utility, which is equivalent to minimizing risk \cite{goel2000minimum,kumar-byrne-2002-minimum,kumar-byrne-2004-minimum}.

MBR decoding consists of two key components: a text generation model $\Pmodel$ and a utility metric $u$. The model estimates the probability of an output $\vy$ given an input sentence $\vx$. The utility metric, $u(\vh, \vy)$, measures the quality of a candidate output $\vh$ with respect to a reference output $\vy$.

Let $\mathcal{Y}$ be a set of all possible sequences.
Given a set of candidate hypotheses $\CandH \subseteq \mathcal{Y}$, MBR decoding selects the hypothesis that maximizes its expected utility:
\begin{equation} 
\vh^{\mathrm{human}} = \argmax_{\vh \in \CandH} \sum_{\vy \in \mathcal{Y}} u(\vh, \vy) \cdot P_{\mathrm{human}}(\vy \mid \vx). 
\end{equation} 
Since the true human probability distribution, $P_\mathrm{human}$, is unknown, MBR approximates it using the model probability $\Pmodel$:
\begin{equation} 
\vh^{\mathrm{model}} = \argmax_{\vh \in \CandH} \sum_{\vy \in \mathcal{Y}} u(\vh, \vy) \cdot \Pmodel(\vy \mid \vx).
\label{eq:mbr} 
\end{equation} 
For simplicity, we denote $\Pmodel$ as $\Pm$ throughout the remainder of this paper unless stated otherwise.

Since integrating over $\mathcal{Y}$ is computationally intractable, Eq.~\eqref{eq:mbr} is typically approximated using a Monte Carlo estimate \cite{eikema-aziz-2022-sampling,farinhas2023empirical}. This is done by sampling a set of reference hypotheses $\RefH$, from the model $\Pm$:
\begin{align} \vh^{\mathrm{MBR}} = \argmax_{\vh \in \CandH} \frac{1}{N} \sum_{\vy \in \RefH} u(\vh, \vy), 
\label{eq:empirical-p} 
\end{align}
where $N = |\RefH|$.

A common practice is to use the same set of hypotheses for both the candidate pool ($\mathcal{H}$) and the reference pool ($\RefH$). We follow the practice in this paper and assume $\CandH = \RefH$.

\subsection{Optimal Transport (OT)}

Optimal Transport (OT; \citealt{peyre2020computational,villani2021topics}) provides a mathematical framework for quantifying the dissimilarity between two distributions. In natural language processing (NLP), OT has been widely used to measure text similarity, often referred to as the Earth Mover’s Distance \cite{pmlr-v37-kusnerb15,zhao-etal-2019-moverscore}. It has been applied in various contexts, including document similarity \cite{pmlr-v37-kusnerb15} and summary evaluation \cite{zhao-etal-2019-moverscore}.

While existing metrics for document-level machine translation have been proposed \cite{vernikos2022}, they are typically designed for tasks where generated documents can be segmented into a fixed sequence of corresponding segments. This limitation makes them unsuitable for scenarios where segment order or count varies across documents.

On the other hand, a key advantage of OT-based metrics is their adaptability to text structure. They can effectively handle variations such as sentence reordering and merging, which frequently occur in machine translation \cite{hovy-gerber-1997-mt,marcu-etal-2000-automatic}. For instance, due to structural differences between Japanese and English, professional translators often restructure paragraphs during translation, significantly altering sentence order and merging content \cite{hovy-gerber-1997-mt}.

In this paper, we consider multiple OT formulations, including linear assignment, Wasserstein distance, and entropic regularized Wasserstein distance.

\paragraph{Linear assignment (LA).}
Linear assignment (LA) is a simple formulation of OT where the cost is computed as a linear sum of the cost of each element \cite{peyre2020computational,villani2021topics}. 
Let $\vh = \{h_1, h_2,\ldots,h_m\}$ and $\vy = \{y_1, y_2,\ldots,y_n\}$ be a set of sentences. Let $p_\vh$ and $p_\vy$ be a probability distribution over a set of elements in $\vh$ and $\vy$.
Let $C$ be a non-negative function representing the cost or the dissimilarity between two sentences.
Using LA, the cost between two sets of sentences is defined as follows:
\begin{align}
\label{eq:la}
&\textrm{LA}_C[p_\vh \| p_\vy] = \nonumber\\
&\inf_{\gamma \in \Gamma_L(\vh, \vy)} \sum_{i \in \{1..m\}} p_\vh(h_i) C(h_i, \gamma(h_i)),
\end{align}
where $\Gamma_L(\vh, \vy)$ is a set of deterministic mappings from $\vh$ to $\vy$ that is injective if $m \leq n$, and subjective if $m \geq n$ (thus bijective if $m=n$).

LA assigns one sentence in a source to exactly one sentence in a target and computes the sum of the cost between them. 
One of the failure cases of this constraint is when aligning texts with and without \textit{merge}.
\begin{quote}
(1)
I like cats and dogs.
\end{quote}
\begin{quote}
(2)
I like cats. I like dogs.
\end{quote}
Because LA cannot distribute weights of a source sentences probabilistically, it has to assign "I like cats and dogs." to only one of either "I like cats." or "I like dogs.".
Thus, although the two texts have very high similarity, its utility computed by LA gets small.

\paragraph{Wasserstein distance (WD).}
The shortcoming of LA is that it is constrained to assign the weight of the source element to a single target element, and thus cannot adapt to the change in disclosure.
Figure~\ref{fig:illustration} shows the example of sentences where this is undesirable.
Wasserstein distance (WD) is a generalization of LA where the source element can divide its weights and assign it to any number of target elements \cite{peyre2020computational,villani2021topics}.
This enables the alignment of the sequence like in Figure~\ref{fig:illustration} where one source sentence is divided into two target sentences.

WD is computed as follows:
\begin{align}
&\textrm{WD}_C[p_\vh \| p_\vy] = \nonumber \\
&\inf_{\gamma \in \Gamma(p_\vh, p_\vy)} \sum_{(i, j) \in m \times n} \gamma(h_i, y_j) \, C(h_i, y_j),    
\end{align}
where $\Gamma(p_\vh, p_\vy)$ is a set of all possible joint distributions $\gamma$ whose marginals are $p_\vh $ and $p_\vy$.

WD is a metric used to quantify the dissimilarity between two probability distributions. Intuitively, it measures the minimum cost required to transform one distribution into the other. This cost is conceptualized as the amount of probability mass that must be moved multiplied by the distance that would be moved.

The advantage of WD over LA is that it can assign multiple reference segments to a single source sentences. In the case of "I like cats and dogs.", one can distribute its weight to both ""I like cats." and "I like dogs.", resulting in high utility score.

\paragraph{Entropic regularized WD (EWD).}
Entropic regularized WD (EWD) is an extension of WD where the KL regularization is enforced to the joint probability distribution $\gamma$ \cite{peyre2020computational,villani2021topics}:
\begin{align}
&\textrm{WD}_C^\epsilon[p_\vh \| p_\vy] \nonumber \\
&=\inf_{\gamma \in \Gamma(p_\vh, p_\vy)} \sum_{(i, j) \in m \times n} \gamma(h_i, y_j) \, C(h_i, y_j) \nonumber \\\
&\;\;\;\;\;\;\;\;\;\;\;\;\;\;\;\;\;\;\;\;\;\;\;+\epsilon \mathrm{KL}[\gamma \| p_\vh \oplus p_\vy],
\end{align}
where $\mathrm{KL}$ represents the KL-divergence between the two probability distributions and $\epsilon$ is a parameter to choose the weight on the KL-divergence term.

Intuitively, EWD is a WD plus the cost of every sentence pair is considered. The KL-divergence term requires the joint distribution $\gamma$ to be spread across $\vy$.
Thus, the KL term is smaller if the two documents are similar to each other overall in addition to the cost of the optimal assignment between pairs of sentences.

EWD is known to be robust under the model uncertainty \cite{azizian2023regularization} and is fast to compute using the Sinkhorn algorithm \cite{cuturi2013sinkhorn}.

\section{\dmbr{}: MBR Decoding using OT}

The performance of MBR decoding relies on the utility function.
However, many of the state-of-the-art utility functions are developed for sentence-level evaluation and are not trained to predict document-level utility.

To this end, we propose \textbf{\dmbr{}}, a variant of MBR decoding that uses WD as the utility function for MBR decoding.
Let the utility function between two sentences be $u_\mathrm{s}$ where we assume its range to be $[0, 1]$.

Let $p_\vh$ be a discrete probability distribution over $\vh = \{\vh_1, \vh_2,...,\vh_m\}$ and so is $p_\vy$ for $\vy = \{\vy_1, \vy_2,...,\vy_n\}$. 
We propose the WD between the two probability distributions to be the utility function of \dmbr{}. Formally, the utility function is defined as follows:
\begin{equation}
    u(\vh, \vy) = 1 - \mathrm{OT}[p_\vh \| p_\vy],
\label{eq:wdmbr}
\end{equation}
where
\begin{equation}
    C(h_i, y_j) = 1 - u_\mathrm{s}(h_i, y_j),
\end{equation}
where $u_\mathrm{s}$ is a utility function to be used for computing the utility of two segments. In the following experiments, we use sentence-level utility functions such as BLEU and MetricX for $u_\mathrm{s}$.
Figure \ref{fig:illustration} shows the illustrative example of how it is computed.

We use a uniform distribution for constructing $p(\vh)$ and $p(\vy)$ so that each sentence is weighted equally: 
\begin{equation}
\label{eq:dmbr}
    p(\vh_i) = \frac{1}{|\vh|}.
\end{equation}
An alternative is to weight a probability proportional to its length so that each token is weighted equally: 
\begin{equation}
\label{eq:dmbrl}
    p_L(\vh_i) = \frac{|\vh_i|}{\sum_{j} |\vh_j|}.
\end{equation}
Incorporating segment informativeness may enhance the performance of our proposed method. This approach may be particularly beneficial in evaluating informal texts where there are large variations in segment informativeness.

The choice of the optimal transport formulation depends on the structure of the language and the objective of the task.
We call the \dmbr{} using AL, WD, and EWD for the OT in Eq.~\eqref{eq:wdmbr} as MBR-AL, \wdmbr{}, and \ewdmbr{}. In particular, we denote these algorithms using Eq.~\eqref{eq:dmbrl} as MBR-AL$_L$, \wdmbrl{}, and \ewdmbrl{}.

The strength of the method is that it can make the best use of the sentence-level utility functions. 
Sentence-level utility functions are meticulously developed by researchers and engineers and they are shown to have a very high correlation with human evaluation \cite{rei-etal-2022-comet,guerreiro-etal-2024-xcomet}. Compared to the effort on sentence-level utility functions, document-level utility functions are less investigated. In fact, many of them are based on using sentence-level utility functions efficiently \cite{liu-etal-2020-multilingual-denoising,vernikos2022}.

\paragraph{Optimizations for \dmbr{} using WD.}
There are several optimizations applicable to speed up \dmbr{} when using WD.
First, because WD (and EWD) are metrics, they are guaranteed to be symmetric:
\begin{equation}
    \forall \vh \forall \vy: u(\vh, \vy) = u(\vy, \vh).
\end{equation}
Also, the value of WD is $1$ when the two distributions are the same. Thus,
\begin{equation}
    \forall \vh: u(\vh, \vh) = 0.
\end{equation}
Therefore, for $N$ documents, one only needs to compute $\frac{N (N - 1)}{2}$ pairs of documents instead of $N^2$ pairs.

Second, because WD is a metric in a finite-dimensional space, it is likely that the matrix of utility scores can be well approximated by a low-rank matrix \cite{drineas2005Nystrom,holodnak2015randomized}. This enables the optimization by a low-rank approximation, significantly reducing the computational complexity \cite{trabelsi2024efficient}.

Third, we can train a document-level utility function by distilling the WD metric, or train the language model directly. There are several studies showing that the performance of LLM can be improved by distilling the output of MBR decoding \cite{yan-etal-2023-bleurt,ramos-etal-2024-aligning,wu2024better,yang-etal-2024-direct,finkelstein2024mbr,guttmann-etal-2024-chasing}.
These approaches require additional training but solve the computational overhead at decoding time.

In the experiments of the paper, we only use the first optimization so that the values are computed exactly.





\section{Experiments}
\label{sec:experiment}

We first evaluate the accuracy of the WD metric for machine translation to assess if it is a metric suitable for a document-level evaluation.
We evaluate the performance of MBR decoding and \dmbr{} on machine translation, document simplification, and dense image captioning tasks.

In all the experiments, we divide the output into a set of sentences for \dmbr{}.
The default sentencizer in spaPy\footnote{\url{https://github.com/explosion/spaCy}} is used for English and German. GiNZA NLP Library\footnote{\url{https://github.com/megagonlabs/ginza}} is used for Japanese as the sentencizer in spaPy is incompatible with Japanese. See Appendix~\ref{apd:hyperparams} for the hyperparameters used for the experiments and Appendix~\ref{apd:reproducibility} for the implementation details. 

\begin{table}
    \centering
    \adjustbox{max width=\columnwidth}{
\begin{tabular}{lllrrr}
\toprule
 &  &  & base & doc & ot (Ours) \\
data & lp & Utility &  &  &  \\
\midrule
\multirow[t]{6}{*}{wmt22} & \multirow[t]{3}{*}{en-de} & sacrebleu & 0.6420 & 0.6783 & 0.7090 \\
 &  & BERTScore & 0.7843 & 0.7670 & 0.8060 \\
 &  & SentBERT & 0.8630 & 0.8313 & 0.8681 \\
\cline{2-6}
 & \multirow[t]{3}{*}{en-ru} & sacrebleu & 0.7861 & 0.7480 & 0.7919 \\
 &  & BERTScore & 0.8106 & 0.7864 & 0.8061 \\
 &  & SentBERT & 0.8030 & 0.8437 & 0.8114 \\
\cline{1-6} \cline{2-6}
\multirow[t]{6}{*}{wmt23} & \multirow[t]{3}{*}{en-de} & sacrebleu & 0.8911 & 0.9165 & 0.9384 \\
 &  & BERTScore & 0.8906 & 0.9212 & 0.9553 \\
 &  & MetricX-23 & 0.9876 & 0.9874 & 0.9753 \\
\cline{2-6}
 & \multirow[t]{3}{*}{he-en} & sacrebleu & 0.8751 & 0.8853 & 0.8716 \\
 &  & BERTScore & 0.8942 & 0.9449 & 0.7138 \\
 &  & MetricX-23 & 0.9357 & 0.9600 & 0.8072 \\
\bottomrule
\end{tabular}
}
    \caption{System level correlation of the metrics with the human evaluation. Base: sentence-level metric, Doc: a document-level metric by \citet{vernikos2022}, WD: a document-level metric using WD.}
    \label{tab:metric}
\end{table}

\subsection{Evaluation of WD Metrics for Machine Translation}
\label{sec:mteval}

We evaluate the accuracy of the WD metric using the Metric Shared Task on WMT22 and WMT23 \cite{freitag-etal-2022-results,freitag-etal-2023-results}.
We compare the correlation of the metrics with the human evaluation.
As a baseline, we compare 1. using the metric to evaluate each segment (Base), and 2. using the document-level evaluation method by \citet{vernikos2022} to evaluate each segment but with a context of the document (Doc).
On computing the score using the WD, we compute the utility of the entire document without using the segmentation provided. Then, we use the average score over a set of documents as the system score. We set $\epsilon=0$ and use a uniform weight (Eq.~\ref{eq:dmbr}) for the probability distribution of WD.

We evaluate BLEU score \cite{bleu-score} using sacrebleu library \cite{post-2018-call}, BERTScore \cite{bert-score} (\texttt{bert-base-multilingual-cased}),\footnote{\url{https://github.com/Tiiiger/bert_score}} and a cosine distance of the sentence embedding model (SentBERT; \texttt{sentence-transformers/all-mpnet-base-v2}), and MetricX-23 (\texttt{google/metricx-23-xl-v2p0}; \citealp{juraska-etal-2023-metricx}). 

Table~\ref{tab:metric} shows the correlation of the metrics with human evaluation. Because MetricX-23 is trained with the WMT22 dataset, we evaluate SentBERT instead.
Overall, we observe the WD metric to be on par with the accuracy of segment-level and document-level evaluation metrics, except for the WMT23 he-en. The authors are not familiar with Hebrew so are not aware of the reason.

Note that the WD metric does not exploit the fact that the sentences of the documents are aligned in the same order. Thus, it uses less information than the baselines.
The result shows that the proposed method is on par with the state-of-the-art metric, while it is a robust metric that is applicable to a wide range of tasks where the disclosure structures can vary.

\begin{table}[t]
    \centering
\begin{tabular}{lrr}
\toprule
MetricX-23 & En-Ja & En-De \\
\midrule
Beam & 61.57 & 79.07 \\
\midrule
MBR (SentBERT) & 56.26 & 78.66 \\
MBR (COMET-22) & 60.55 & 80.92 \\
MBR (SFR-2) & 57.38 & 76.88 \\
MBR (MetricX) & 68.81 & 82.02 \\
\midrule
MBR-LA (MetricX) & 70.01 & 80.77 \\
MBR-LA$_L$ (MetricX) & 71.01 & 83.13 \\
\ewdmbr{} (MetricX) & 68.07 & \textbf{83.77} \\
\ewdmbrl{} (MetricX) & 70.67 & 83.24 \\
\wdmbr{} (MetricX) & \textbf{75.29} & 83.40 \\
\wdmbrl{} (MetricX) & 72.38 & 83.24 \\
\bottomrule
\end{tabular}
    \caption{Comparison of MBR-OT methods with LA, WD, and EWD. The evaluation metric is MetricX-23-XXL with EWD.}
    \label{tab:mt-ot}
\end{table}

\begin{figure}[tb]
    \centering
    \subfloat[WMT24 En-Ja]{\includegraphics[width=0.9\columnwidth]{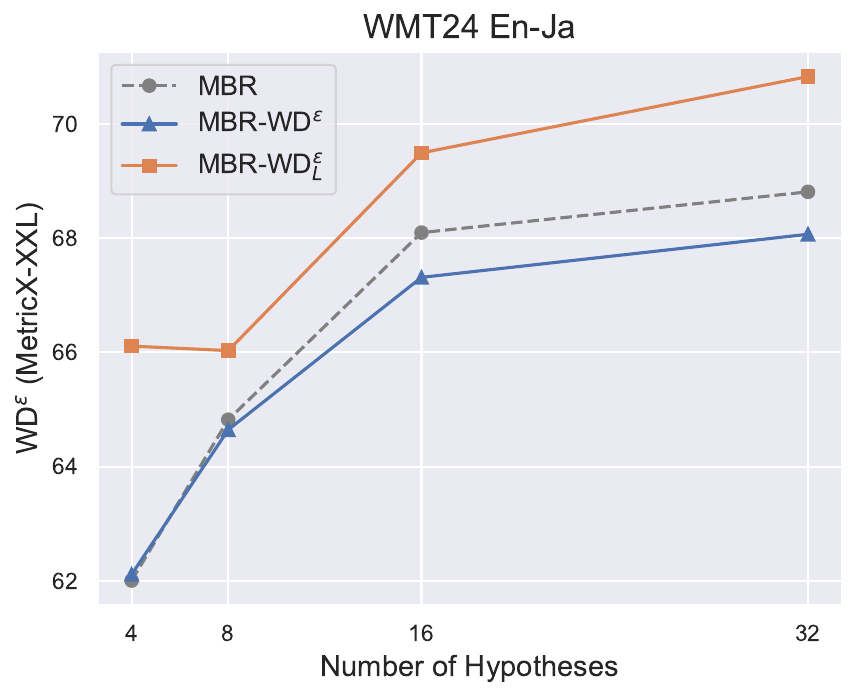}}\\
    \subfloat[WMT24 En-De]{\includegraphics[width=0.9\columnwidth]{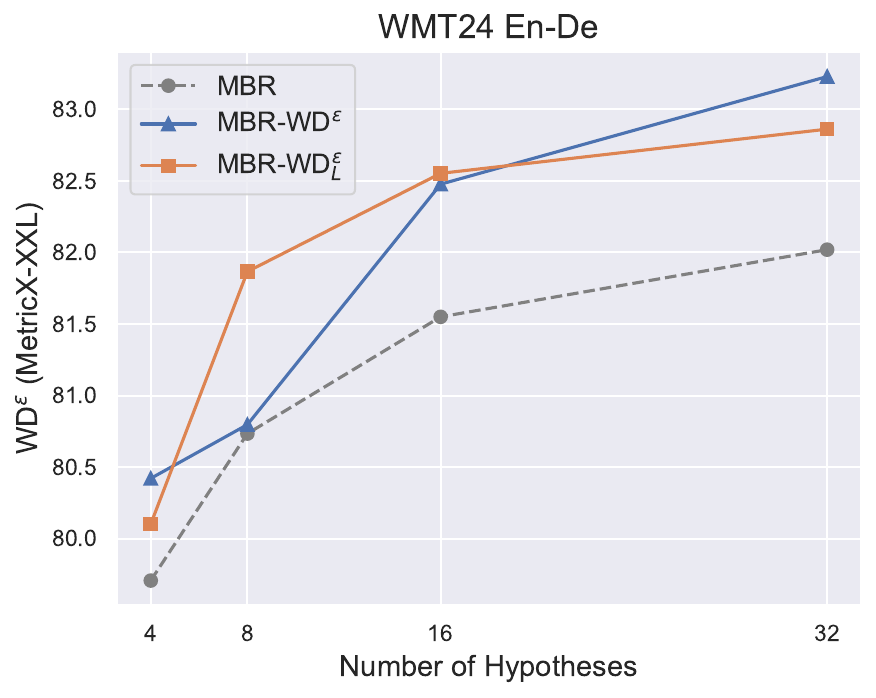}}
    \caption{Evaluation of \dmbr{} on document-level machine translation tasks. WD metric with MetricX-23 as a sentence-level utility function is used as the evaluation metric.}
    \label{fig:mt-results}
\end{figure}

\begin{table}
    \centering
\begin{tabular}{lrr}
\toprule
COMET-22 & En-Ja & En-De \\
 & CALM2-DPO & EuroLLM \\
\midrule
MBR & 	79.65 & 83.03 \\
\ewdmbr{} & 83.54 & \textbf{85.13} \\
\ewdmbrl{} & \textbf{84.08} & 84.25 \\
\bottomrule
\end{tabular}
    \caption{Evaluation of \ewdmbr{} on WMT24 with 32 samples on LLMs specifically trained for the languages (\texttt{cyberagent/calm2-7b-chat-dpo-experimental} and \texttt{utter-project/EuroLLM-1.7B-Instruct}). The evaluation metric is MetricX-23-XXL with EWD.}
    \label{tab:mt-others}
\end{table}

\begin{table}
    \centering
\begin{tabular}{lrr}
\toprule
BLEU & En-Ja & En-De \\
\midrule
Beam & 7.75 & 17.44 \\
MBR & 10.01 & 18.42 \\
\ewdmbr{} & 9.66 & 18.26 \\
\ewdmbrl{} & 9.82 & 18.42 \\
\bottomrule
\end{tabular}
    \caption{BLEU scores of the \ewdmbr{} and MBR on WMT24 with 32 samples. \texttt{mecab-python3} library \cite{kudo-etal-2004-applying} is used for tokenizing Japanese texts. Note that the lexical metric is shown to have little correlation with human evaluation. The values are reported for reference.}
    \label{tab:mt-lex}
\end{table}

\begin{table*}
    \centering
    \adjustbox{max width=\textwidth}{
    \begin{tabular}{lrrrrrrrrr}
    \toprule
        Dataset & \multicolumn{3}{c}{MBR} & \multicolumn{3}{c}{\ewdmbr{}} & \multicolumn{3}{c}{\ewdmbrl{}} \\
        & |C| & |S| & |C|/|S| & |C| & |S| & |C|/|S| & |C| & |S| & |C|/|S| \\
        \cmidrule(lr){2-4} \cmidrule(lr){5-7} \cmidrule(lr){8-10} 
        WMT24 En-Ja &  713.6 & 28.4 & 25.2 & 633.9 & 29.4 & 21.6 & 625.0 & 29.6 & 21.1 \\
        WMT24 En-De &  1435.0 & 12.3 & 116.4 & 1366.9 & 11.8 & 115.3 & 1366.0 & 11.6 & 117.4 \\ 
        JADOS & 338.7 & 7.8 & 43.6 & 317.9 & 7.3 & 43.5 & 319.3 & 7.2 & 44.2  \\
        CNNDM & 488.5 & 3.0 & 162.8 & 489.7 & 2.9 & 168.9 & 471.1 & 2.9 & 165.3 \\
        PP-cc12m & 660.5 & 5.9 & 111.4 & 589.0 & 6.9 & 85.5 & 625.9 & 7.2 & 86.7 \\
        PP-commonpool & 565.9 & 5.1 & 110.3 & 517.8 & 6.1 & 84.9 & 508.0 & 5.9 & 86.5 \\
        PP-redcaps & 710.5 & 6.5 & 109.3 & 572.7 & 6.8 & 84.5 & 586.7 & 6.5 & 90.1 \\
    \bottomrule
    \end{tabular}
    }
    \caption{The average number of characters (|C|), sentences (|S|), and the average number of characters per sentence (|C|/|S|).}
    \label{tab:length}
\end{table*}

\subsection{Document-Level Translation}
\label{sec:mt}

The use of LLMs for document-level translation is shown to be effective due to their ability to comprehend long-context texts \cite{wang-etal-2023-document-level}.

We use WMT24 En-De and En-Ja language pairs for the evaluation \cite{kocmi-etal-2024-findings}.
We generate 32 samples using Llama-3.1 (\texttt{meta-llama/Llama-3.1-8B-Instruct}) as a language model \cite{dubey2024llama}. See Appendix \ref{apd:prompt} for the prompt used.

We use MetricX-23 (\texttt{google/metricx\--23\--xl\--v2p0}) as the utility function \cite{juraska-etal-2023-metricx}.
Because the output from the MetricX-23 models is a score in the range $[0, 25]$ where lower is better, we inverted and rescaled it to $[0, 1]$.
For an evaluation, we use MetricX-23 with EWD ($\epsilon=0.1$) to evaluate document-level text generation. Given that MetricX-23 with EWD shows a high correlation with human evaluation in Table~\ref{tab:metric}, we foresee it to be a valid metric for the task. 
In addition to using MetricX as a utility function, we also evaluate using SentBERT, COMET-22, and SFR-2 (\texttt{Salesforce/SFR-Embedding-2\_R}) as a reference.
We use a bigger MetricX-23 model (\texttt{google/metricx-23-xxl-v2p0}) to alleviate the overfitting problem of MBR decoding \cite{kovacs-etal-2024-mitigating}.
We additionally evaluate with COMET-22 (\texttt{Unbabel/wmt22-comet-da}; \citealt{rei-etal-2022-comet}) in Appendix~\ref{apd:mt-comet}.

\paragraph{Comparison of LA, WD, and EWD.} 
Table~\ref{tab:mt-ot} shows the performance of \dmbr{} using LA, WD, and EWD as the formulation of the optimal transport.
We observe that \ewdmbr{} and \wdmbr{} outperform the baselines.

We observe the BLEU scores of En-Ja are low compared to the state-of-the-art models (Table~\ref{tab:mt-lex}). 
Llama-3 tends to generate a shorter summary of the original English document rather than a precise translation. The average length of the generated texts is 642.34 characters, whereas the reference texts average 860.61 characters. In several cases, the generated outputs are less than one-third the length of the reference translations.
Table~\ref{tab:length} summarizes the number of characters, sentences, and the average number of characters per sentences generated by MBR and \dmbr{}.

For the rest of the paper, we conduct experiments using EWD with $\epsilon=0.10$, as its performance is only marginal below WD, and EWD is known to be robust to model noises.

\paragraph{Evaluation of \dmbr{}.}
Figure~\ref{fig:mt-results} shows the performance of the algorithms on Llama-3.1 and Table~\ref{tab:mt-others} on LLMs specifically trained for the target languages.
The result shows that the performance of \dmbr{} is outperforming the baselines in both language pairs.

\begin{table}[t]
    \centering
\begin{tabular}{lr}
\toprule
 & jReadability \\
\midrule
Beam & 4.46 \\
MBR & 3.30 \\
\ewdmbr{} & \textbf{3.25} \\
\ewdmbrl{} & \textbf{3.25} \\
\bottomrule
\end{tabular}
    \caption{Evaluation of readability of the generated text using jReadability \cite{hasebe2015introducing}. The lower score shows better readability.}
    \label{tab:jreadability}
\end{table}

\subsection{Document-Level Simplification}
\label{sec:simplification}

Document-level simplification task is a combination of document summarization and a text simplification task \cite{sun-etal-2021-document,blinova-etal-2023-simsum}. The goal is to generate a short and easily readable summary of the given document so that the information is accessible to children and those who are learning the language.
We use the first 300 entries of the JADOS dataset for the task \cite{nagai-etal-2024-document}.
The dataset contains articles in Japanese and their summaries, which were manually written by native Japanese speakers.
We use the Wikipedia subset of the JADOS dataset as it is open-sourced.

We generate 32 samples using Llama-3.1 as a language model.
We use the SentBERT as a utility function.
D-SARI \cite{sun-etal-2021-document,blinova-etal-2023-simsum} is used as the document-level evaluation metric following \citet{nagai-etal-2024-document}.

\begin{figure}[tb]
    \centering
    \subfloat[JADOS]{\includegraphics[width=0.9\columnwidth]{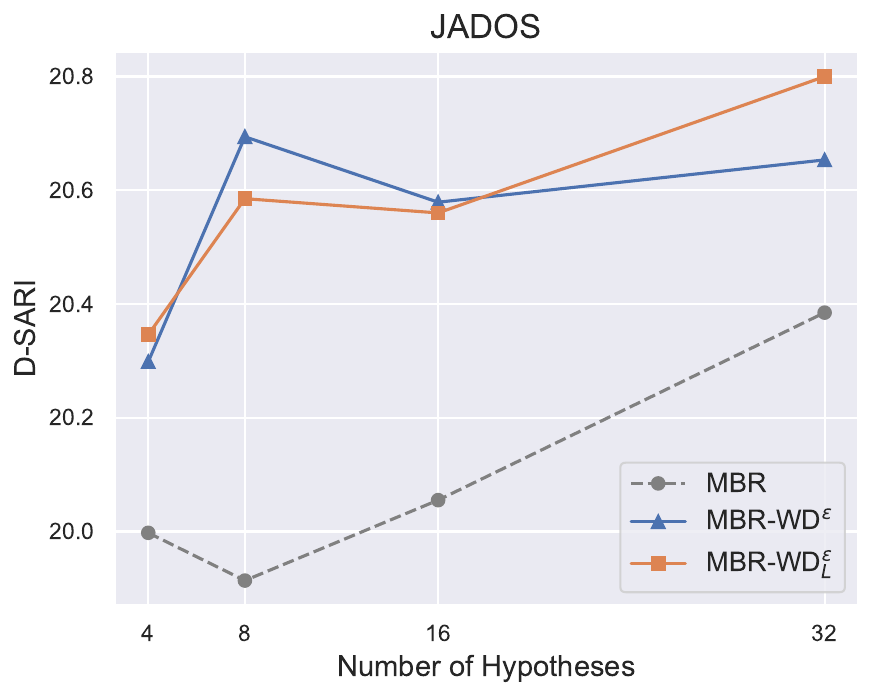}\label{fig:jados}}\\
    \subfloat[CNNDM]{\includegraphics[width=0.9\columnwidth]{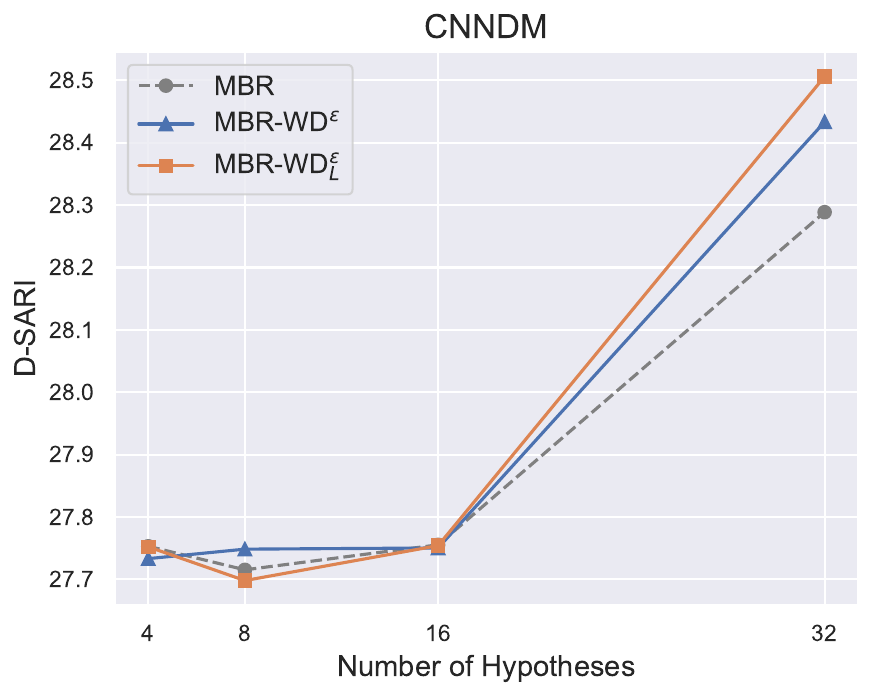}\label{fig:cnndm}}
    \caption{Evaluation of \dmbr{} on document-level summarization and simplification tasks.}
    \label{fig:sum-results}
\end{figure}

Figure \ref{fig:jados} shows the D-SARI scores of the decoding algorithms with varying number of samples. Overall, we observe \dmbr{} outperforming the baselines.
We additionally evaluate the readability of the texts using JReadability \cite{hasebe2015introducing} in Table~\ref{tab:jreadability} which shows that the readability of \dmbr{} is on par with MBR. Table~\ref{tab:sum-lex} shows the ROUGE scores as a reference.

\begin{table}
    \centering
\begin{tabular}{lrr}
\toprule
 & CNNDM & JADOS \\
\midrule
Beam & 16.32 & 16.39 \\
MBR & 17.58 & 25.39 \\
\ewdmbr{} & 18.23 & 26.63 \\
\ewdmbrl{} & 18.88 & 26.56 \\
\bottomrule
\end{tabular}
    \caption{ROUGE scores of the \ewdmbr{} and MBR on CNNDM and JADOS with 32 samples. Note that the lexical metric is shown to have little correlation with human evaluation. The values are reported for reference.}
    \label{tab:sum-lex}
\end{table}

\subsection{Document-Level Summarization}
\label{sec:cnndm}
Although the proposed method is targeted for document-level text generation tasks, it is applicable to generating short paragraphs with a couple of sentences. We evaluate the performance of \dmbr{} on the first 300 entries of the CNNDM dataset where the output is around 2 to 4 sentences.
CNNDM is a summarization task where the goal is to generate a short summary of the given news article.
We use Llama-3.1 as a language model and SentBERT as a utility function.
The result shows that the present approach has a positive impact even when the number of segments in the output is relatively small.

\begin{table}
    \centering
\begin{tabular}{lrrr}
\toprule
 & cc12m & commonpool & redcaps\\
\midrule
Beam & 28.26 & 24.77 & 27.57 \\
MBR & 28.38 & 25.53 & 29.27 \\
\ewdmbr{} & 27.99 & 25.20 & 29.03 \\
\ewdmbrl{} & 28.28 & 24.33 & 28.70 \\
\bottomrule
\end{tabular}
    \caption{Evaluation of the \ewdmbr{} and MBR using METEOR with 32 samples. Note that the lexical metric is shown to have little correlation with human evaluation. The values are reported for reference.}
    \label{tab:captioning-lex}
\end{table}

\begin{figure}[tb]
    \centering
    \subfloat[PP-cc12m]{\includegraphics[width=0.73\columnwidth]{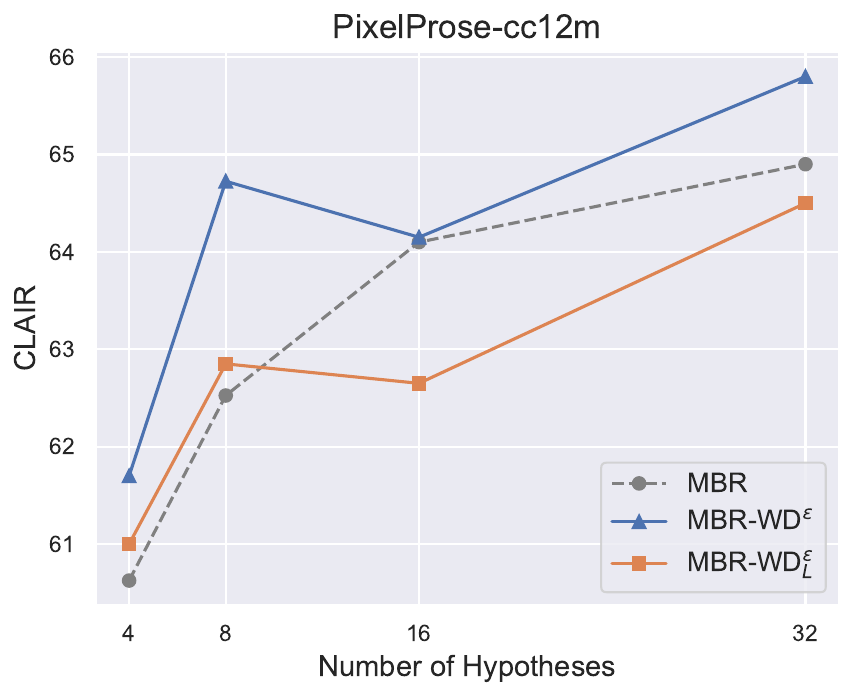}}\\
    \subfloat[PP-commonpool]{\includegraphics[width=0.73\columnwidth]{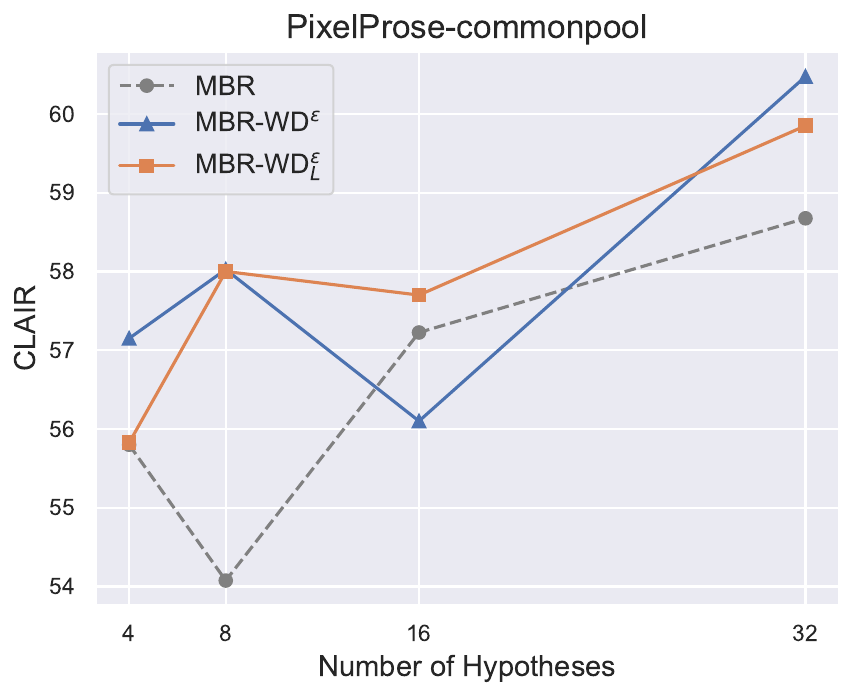}}\\
    \subfloat[PP-redcaps]{\includegraphics[width=0.73\columnwidth]{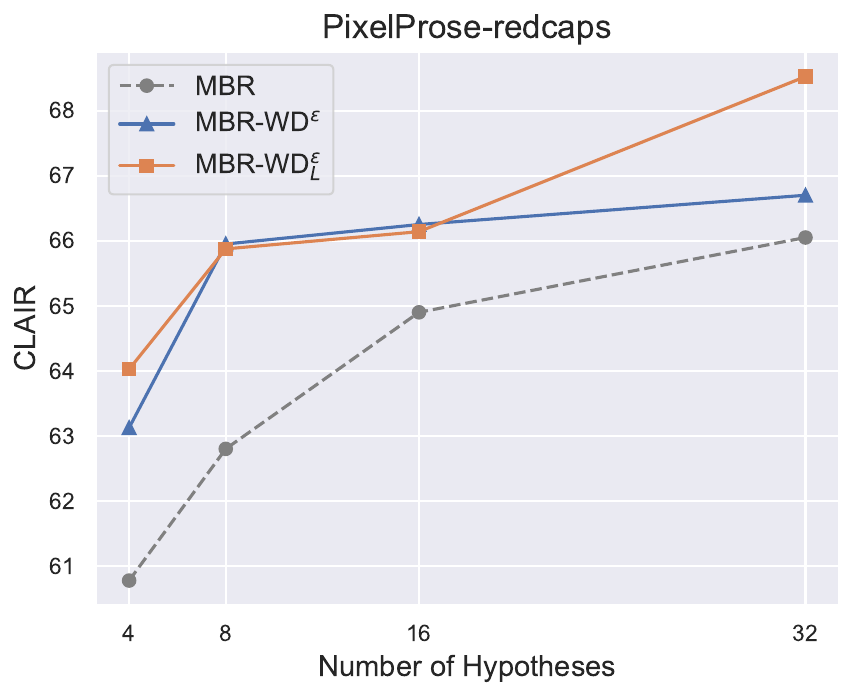}}
    \caption{Evaluation of \dmbr{} on dense image captioning tasks.}
    \label{fig:captioning-results}
\end{figure}

\subsection{Dense Image Captioning}
\label{sec:captioning}

The task of dense image captioning is to generate a caption of an image containing types, attributes, and counts of objects in an image, in addition to spatial relations between objects, the presence of text, various broad image categorizations, etc \cite{Johnson_2016_CVPR,krishna2017visual,Urbanek_2024_CVPR,singla2024pixels}.

Unlike a traditional image captioning task, the caption wants to contain as much information about the image as possible.

We use the PixelProse dataset for the evaluation \cite{singla2024pixels}.
PixelProse is a collection of images crawled web pages with curation to filter out harmful images and images that can violate privacy.
We use 200 images from each of the three subsets of the dataset (cc12m, commonpool, and redcaps).
We use PaliGemma-2 10B \cite{PaliGemma} fine-tuned on DOCCI dataset \cite{docci2024} to generate 32 captions for each image (\texttt{google/paligemma2-10b-ft-docci-448}).
We use CLIPText as a utility function (\texttt{openai/clip-vit-large-patch14}; \citealp{qin-etal-2023-cliptext}).
CLAIR \cite{chan-etal-2023-clair} is used as the metric to evaluate the outputs. CLAIR uses GPT-4 as a judge to compute the utility of the caption. 

Figure~\ref{fig:captioning-results} shows the performance of the algorithms. Overall, the performance of \wdmbr{} outperforms MBR decoding in all three subsets, showing that it consistently achieves high quality captioning in a wide range of images.
We observe little differences between the algorithms on METEOR \cite{banerjee-lavie-2005-meteor} scores (Table~\ref{tab:captioning-lex}).




\section{Related Work}
\label{sec:related}
\paragraph{MBR decoding.}

The performance of MBR decoding is known to be dependent on the quality of the utility function \cite{fernandes-etal-2022-quality,freitag-etal-2022-high,kovacs-etal-2024-mitigating}.
One of the problems of MBR decoding is its computational complexity. Several methods have been proposed to reduce the computational complexity of MBR decoding \cite{cheng-vlachos-2023-faster,jinnai2024hyperparameterfree,trabelsi2024efficient}, many of which are also applicable to \dmbr{}. 

The sampling strategy is an important aspect of MBR decoding. Epsilon sampling \cite{hewitt-etal-2022-truncation} is known to be an effective choice for MBR \cite{freitag-etal-2023-epsilon,jinnai2023modelbased}. We use epsilon sampling in this paper following their work.


\paragraph{Document-level text generation tasks.}

Improvement to the decoding algorithm is not the only solution to solve document-level text generation tasks.

\citet{gpt-mt-2023} show that few-shot learning is an effective strategy to generate high quality texts on document-level translation tasks using GPT.
\citet{briakou-etal-2024-translating} propose a method to translate a document by decomposing the translation process into steps consisting of pre-translation research, drafting, refining, and proofreading.
\citet{jiang-etal-2022-blonde} proposes a metric specific to the evaluation of document-level machine translation tasks. 

\citet{blinova-etal-2023-simsum} proposes to use both the explicit summarization model and the simplification model to generate the final output.
\citet{cripwell-etal-2023-document} proposes a method to plan a simplification as a sequence of four simplification operations (copy, rephrase, split, or delete).

Because the objective of dense image captioning is to extract as much information from the given image, many of the studies have proposed methods to extract information from the image effectively. 
The importance of dense image captioning is recognized by the study by \citet{krishna2017visual} is one of the first studies to show the use of rich annotation data for images.
\citet{hao2024fullanno} combines a detection model, a vision-language model, an OCR model, and other CV tools to generate dense image captions.




While these methods achieve domain-specific optimization to improve the performance of the given task, our contribution is to present a general-purpose decoding algorithm that is independent of the domain, except for its length. 


\section{Conclusions}

We evaluate the performance of \dmbr{}, a variant of MBR decoding using optimal transport distance to compute the document-level utility using a segment-level utility function.
Compared to the prior work on document-level metric for machine translation \cite{vernikos2022}, the advantage of \dmbr{} is that it is naturally applicable to tasks where there are many options for disclosure structures in the output text. WD can adapt to the reordering, merging, and separation of disclosures, making it applicable to many tasks without engineering task-dependent optimization.

The empirical result shows that \dmbr{} outperforms MBR decoding in document-level text generation tasks including machine translation, text simplification, and dense image captioning.

\section{Limitations}

\dmbr{} requires additional computational overhead on MBR decoding which is already known to be computationally intensive. In our experiments using NVIDIA A100 GPUs with 80 GB VRAM, \dmbr{} is roughly more than four times slower than MBR with the same number of samples.
We foresee it to be used combined with efficient MBR methods \cite{cheng-vlachos-2023-faster,jinnai2024hyperparameterfree} in practice.
Also note that approximation algorithms to compute optimal transport distance faster have been proposed \cite{NIPS2017_491442df,pmlr-v80-dvurechensky18a,JMLR:v20:18-079}, which may enable us to compute \dmbr{} faster.

Our study is limited to simple formulations of OT where the ordering of the segments are ignored.
This may result in cases where the ordering of the segments matters. For example, the meaning of the pronoun is dependent on the ordering:
\begin{quote}
(1) 
Bob likes cats.
Charles likes dogs.
*He* always takes pictures of them.    
\end{quote}

\begin{quote}
(2)
Charles likes dogs.
Bob likes cats.
*He* always takes pictures of them.    
\end{quote}
In the first text, “He” refers to Charles and “them” refers to dogs, but in the second text, they refers to Bob and cats. Therefore, the third sentence has semantically different meanings in the context. However, the formulations we present in the paper cannot distinguish such a difference.
This suggests that the use of additional context proposed by \citet{vernikos2022} may be also beneficial to the OT utility functions. 
Alternative approach is to use a more sophisticted OT formulation such as Fused Gromov-Wasserstein \cite{pmlr-v97-titouan19a} to take into account the structures of the segments.

We consider the weight of the segments to be uniform or propotional to its length (Eq.~\ref{eq:dmbr} and \ref{eq:dmbrl}). Evaluation of other approaches such as using probability mass or entropy would be future work.

We focus on directed text generation tasks.
Open-ended text generation tasks such as story generation is an interesting future work.


The evaluation is limited to moderately long documents with a couple of paragraphs. The evaluation of the method for much longer text generation tasks \cite{liang-etal-2023-open,zhang-etal-2024-bench,hsieh2024ruler} is future work.
For generating much longer texts, we require the segmentation of paragraphs instead of sentences to align larger semantic blocks. How to build a hierarchy of segmented texts from a long text is not a trivial problem that we need to investigate for applying \dmbr{} to these tasks.

As mentioned in Section~\ref{sec:related}, there are many methods proposed for each specific document-level text generation task. For example, several methods have been proposed for document-level machine translation \cite{voita-etal-2019-context,kudo-etal-2024-document}.
Evaluation of \dmbr{} combined with task-dependent optimizations is future work.

The study depends on automated metrics. Although the metrics we used in the study are known to have a high correlation with human evaluation, they are not flawless.
Human evaluation is desirable for evaluation.

\section{Impact Statements}

This work was conducted using existing, publicly available, including WMT datasets, JADOS, PixelProse, and CNNDM. The datasets are constructed as benchmarks for research use in the research community (Table~\ref{tab:links}). 

As for MBR, \dmbr{} is not designed to prevent the system from generating toxic texts. Thus, it requires a countermeasure besides \dmbr{}, such as language model alignment \cite{NEURIPS2022_b1efde53,eisenstein2023helping,rafailov2023direct}.

\section*{Acknowledgments}
We thank the anonymous reviewers for their insightful comments on the manuscript, which enabled us to conduct a more fine-grained analysis of the experiments.
We would like to thank our colleagues at CyberAgent AI Lab, Kenshi Abe, Kaito Ariu, Tetsuro Morimura, Mitsuki Sakamoto, Yuma Fujimoto, and Ryota Mitsuhashi for their encouragement and for the helpful discussions during the early stages of this work.

\bibliography{anthology,ms,ms2,doc}
\appendix

\section{Additional Evaluation using COMET-22}
\label{apd:mt-comet}
Using the same evaluation metric as the utility function that MBR decoding uses is known to cause a bias in the evaluation. Prior work has shown that the metric overestimates the performance of MBR decoding using the evaluation metric \cite{kovacs-etal-2024-mitigating}.

Thus, the evaluation on Section~\ref{sec:mt} may exhibit bias. To this end, we additionally evaluate the outputs using COMET-22 (\texttt{Unbabel/wmt22-comet-da}; \citealt{rei-etal-2022-comet}) as a verification of the evaluation.

Tables~\ref{tab:mt-ot-comet} and \ref{tab:mt-comet-others} show the comparison of MBR-OT methods using varying OT formulations. Overall, we observe that MBR-WD outperforms the baselines.

\begin{figure}[tb]
    \centering
    \subfloat[WMT24 En-Ja]{\includegraphics[width=0.95\columnwidth]{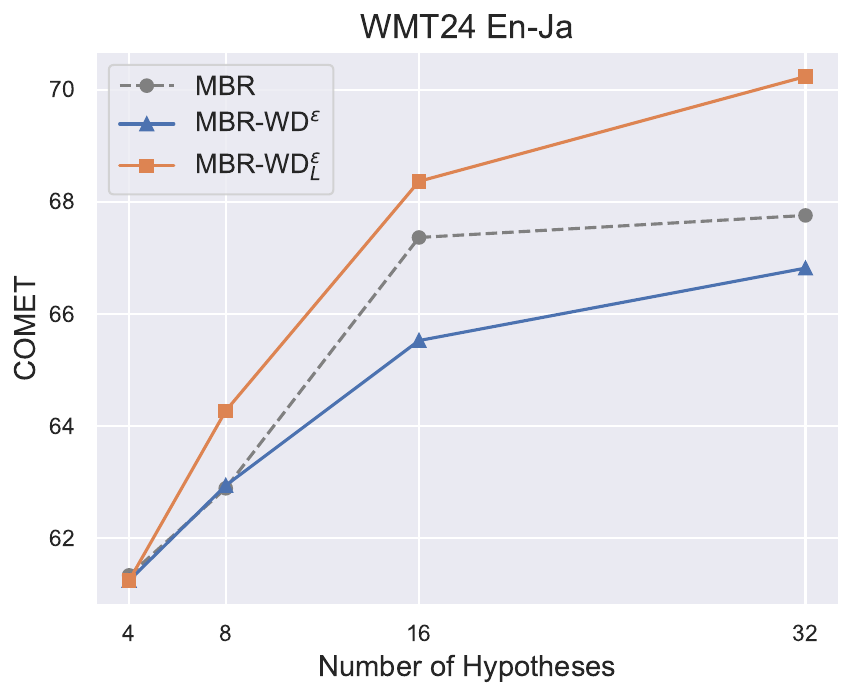}}\\
    \subfloat[WMT24 En-De]{\includegraphics[width=0.95\columnwidth]{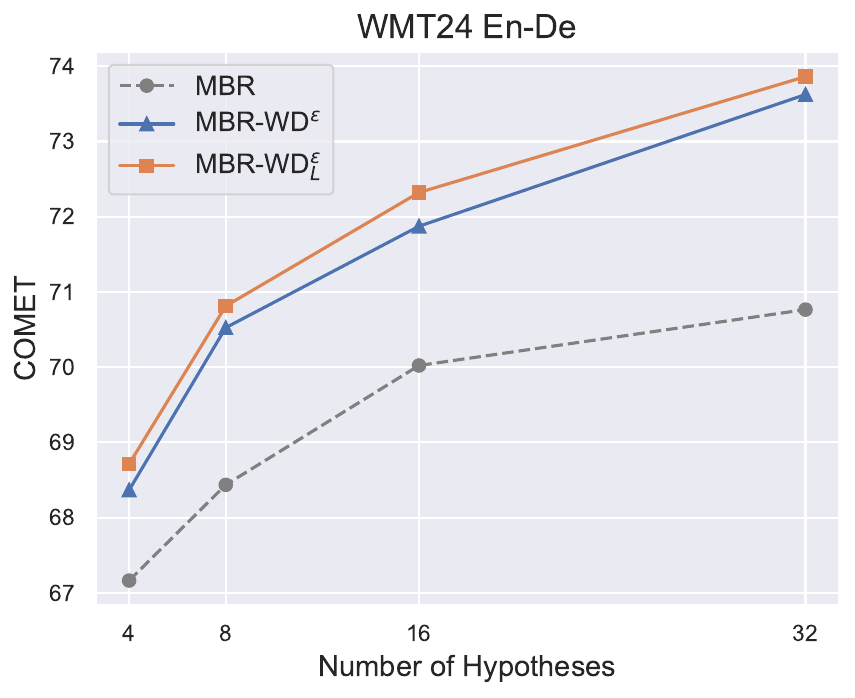}}
    \caption{Evaluation of \dmbr{} on document-level machine translation tasks using COMET-22. Llama-3.1 is used as the text generation model.}
    \label{fig:mt-comet-results}
\end{figure}

Figure~\ref{fig:mt-comet-results} shows the performance of MBR-WD with varying numbers of hypotheses. Overall, the same qualitative results are observed as in Figure~\ref{fig:mt-results}.


\section{Analysis of \dmbr{} on WMT24 En-Ja}

To further investigate the behavior of \dmbr{}, we computed the average scores for each domain in the WMT24 En-Ja task (Table~\ref{tab:domains}). Note that the number of documents is uneven (news = 17, speech = 111, literary = 8, and social = 34). The performance of \wdmbr{} (MetricX) is better than MBR (MetricX) in all four categories. Interestingly, in the social domain, where MBR drops the score compared to the other domains, \wdmbr{} achieves a score on par with the other domains. We speculate that this is because documents in social domains are sourced from SNS (Mastodon) \cite{kocmi-etal-2024-findings} and are less structured than those in news domains sourced from online news sites. OT may not offer a strong advantage in the news domain as the document structure is relatively fixed.

Note that the analysis is based on a small number of samples, so further investigation is needed for more fine-grained analysis.

\begin{table}
    \centering
\begin{tabular}{lrr}
\toprule
COMET-22 & En-Ja & En-De \\
\midrule
Beam & 58.14 & 67.19 \\
MBR (SFR2) & 53.98 & 61.72 \\
\midrule
MBR (MetricX) & 67.76 & 70.77 \\
MBR-LA (MetricX) & 64.12 & 72.56 \\
MBR-LA$_L$ (MetricX) & 64.56 & 73.17 \\
\wdmbr{} (MetricX) & 66.82 & 72.70 \\
\wdmbrl{} (MetricX) & \textbf{70.46} & 73.13 \\
\ewdmbr{} (MetricX) & 66.82 & 73.62 \\
\ewdmbrl{} (MetricX) & 70.23 & \textbf{73.86} \\
\bottomrule
\end{tabular}
    \caption{Comparison of MBR-OT methods with LA, WD, and EWD using COMET-22. Llama-3.1 is used as the text generation model.}
    \label{tab:mt-ot-comet}
\end{table}

\begin{table}
    \centering
\begin{tabular}{lrr}
\toprule
COMET-22 & En-Ja & En-De \\
 & CALM2-DPO & EuroLLM \\
\midrule
MBR & 83.26 & 75.44 \\
\ewdmbr{} & 83.06 & 76.27 \\
\ewdmbrl{} & \textbf{83.55} & \textbf{76.35} \\
\bottomrule
\end{tabular}
    \caption{COMET-22 scores of the \dmbr{} and MBR on WMT24 with 32 samples on LLMs specifically trained for the languages.}
    \label{tab:mt-comet-others}
\end{table}

\begin{table*}
    \centering
\begin{tabular}{lrrrrr}
\toprule
Category & Overall & News & Speech & Literary & Social \\
\midrule\midrule
Beam & 61.57 & 63.22 & 60.55 & 56.01 & 65.37 \\\midrule
MBR (SentBERT) & 56.26 & 72.46 & 54.38 & 52.51 & 60.35 \\
MBR (COMET-22) & 60.55 & \textbf{77.06} & 58.15 & 56.17 & 61.16 \\
MBR (SFR-2) & 57.38 & 70.40 & 54.35 & 61.23 & 59.86 \\
MBR (MetricX) & 68.81 & 73.32 & 71.14 & 68.48 & 59.03 \\\midrule
MBR-LA (MetricX) & 70.01 & 77.34 & 72.64 & \textbf{73.62} & 66.93 \\
\ewdmbr{} (MetricX) & 68.07 & 74.64 & 70.41 & 58.77 & 59.31 \\
\wdmbr{} (MetricX) & \textbf{75.29} & 77.00 & \textbf{75.94} & 74.96 & \textbf{72.40} \\
\bottomrule
\end{tabular}
    \caption{Evaluation of the MetricX scores on the WMT'2024 En-Ja. The number of documents is news=17, speech=111, literary=8, and social=34.}
    \label{tab:domains}
\end{table*}

\section{Generation Examples on WMT24 En-Ja}

Below are the generation examples of the Llama-3.1 model by random sampling.
The source English document is the following. 

\paragraph{Source document (Abbreviated)}
\begin{quote}
          Siso's depictions of land, water center new gallery exhibition
          "People Swimming in the Swimming Pool" from 2022 is one Vicente Siso artwork that will display at Tierra del Sol Gallery beginning Jan. 13. (photo courtesy of Vicente Siso)
          Tierra del Sol is pleased to present "Vicente Siso: Memories of the Land and Water" at the new gallery location in West Hollywood. Siso has been an artist in the Studio Arts Program since 2012, this marks his debut solo exhibition. 
          ...
          "Vicente Siso: Memories of the Land and Water" opens on Saturday, Jan. 13, with a reception from 6-8 p.m. The exhibition is on view through Sunday, March 3.
          The Tierra del Sol Gallery is located at 7414 Santa Monica Blvd. For information, visit tierradelsolgallery.org.
\end{quote}

\paragraph{Generation example 1}
\begin{quote}
\jp{
Sisoの作品、「2022年に製作された「Swimming Poolの人間たち」」が、2023年1月13日からTierra del Sol Galleryで展示されることになります。この展示は、西ハリウッドの新しいギャラリーで開催される「Vicente Siso:ランドと水の記憶」というタイトルのグループ展です。
...
「Vicente Siso:ランドと水の記憶」は、2023年1月13日に開幕し、6-8 p.m.の受け入れを含み、3月3日まで展示します。ティエラデルソル・ギャラリーは、サンタモニカブラビュー7414号にあります。情報は、tierradelsolgallery.orgにアクセスしてください。    
}
\end{quote}

\paragraph{Generation example 2}
\begin{quote}
\jp{
2022年の作品「People Swimming in the Swimming Pool」が、西海岸の Tierra del Sol ギャラリーで展示されるVICENTE SISOの作品です。Tierra del Sol ギャラリーは、ロサンゼルス西部の新しいギャラリーで、VICENTE SISOの個展「Memories of the Land and Water」を開催します。

...

VICENTE SISOの個展「Memories of the Land and Water」は、1月13日の土曜日6-8pmに開催され、3月3日の日曜日まで展示されます。Tierra del Sol ギャラリーは、サンタモニカ・ブールバード7414番地に所在しており、tierradelsolgallery.orgで情報を閲覧できます。
}
\end{quote}

\paragraph{Generation example 3}
\begin{quote}
\jp{
シソ・ソロ展「Vicente Siso：Land and Waterのメモリー」

新しいギャラリー、ティエラ・デル・ソル・ギャラリーにおいて、2022年の「People Swimming in the Swimming Pool」を始まり、1月13日から出展します。（写真：Vicente Siso）
...
「Vicente Siso：Land and Waterのメモリー」は、1月13日(土)6-8pmのオープン・リセプションに始まり、3月3日(日)まで開催されます。
ティエラ・デル・ソル・ギャラリーは、サンタ・モニカ・ブールバード7414番地にあります。詳細はtierradelsolgallery.orgを参照してください。    
}
\end{quote}

The “\jp{。}” character serves as the period in Japanese, so most texts are segmented based on this character. In the examples, the bolded text highlights segments where the structure differs from the rest of the generated outputs. In the second example, two sentences from the original English document are merged into a single sentence. In the third example, the first sentence of the document appears to be interpreted as the title of the article.

Although the document-level MT is a directed text generation task, the generated texts from the LLM may have various structures.

\section{Prompts for Text Generation}
\label{apd:prompt}
We use the following prompt for the machine translation tasks in Section \ref{sec:mt}.

\begin{quote}
Translate the following paragraph from English to [[LANGUAGE]].

[[QUESTION]]

Translate this paragraph from English to [[LANGUAGE]].
\end{quote}

We use the following prompt for the JADOS dataset (Section \ref{sec:simplification}).

\begin{quote}
\jp{
以下のWikipediaの記事を、小学生でも分かるやさしい日本語で要約してください。\\
\\
記事: [[QUESTION]]\\
Q: この記事をやさしい日本語で要約してください。\\
A: 
}
\end{quote}

The text below is the above prompt translated into English.

\begin{quote}
Please summarize the following Wikipedia article in simple Japanese that even an elementary school student can understand.\\
\\
Article: [[QUESTION]]\\
Q: Please summarize this article in simple Japanese.\\
A:
\end{quote}

For the dense image captioning (Section \ref{sec:captioning}), we use the same prompt as \citealt{singla2024pixels}:
\begin{quote}
    Describe the image in detail. Please specify any objects within the image, backgrounds, scenery, interactions, and gestures or poses. If they are multiple of any object, please specify how many and where they are. If any text is present in the image, mention where it is, and the font. Describe the text in detail with quotation marks. For example, if the image has text, Merry Christmas, write it down as “Merry Christmas”. Describe the style of the image. If there are people or characters in the image, what emotions are they conveying? Identify the style of the image, and describe it as well. Please keep your descriptions factual and terse but complete. The description should be purely factual, with no subjective speculation. Make sure to include the style of the image, for example cartoon, photograph, 3d render etc.
\end{quote}

\section{Hyperparameters}
\label{apd:hyperparams}

Table~\ref{tab:hyperparams} shows the hyperparameters used for text generation. We use epsilon sampling \cite{hewitt-etal-2022-truncation} for all the experiments as it is shown to be effective for generating samples for MBR decoding \cite{freitag-etal-2023-epsilon,jinnai2023modelbased}.

Table~\ref{tab:hyperparams-gpt4} shows the hyperparameters we use for GPT-4 evaluation and for CLAIR in Section~\ref{sec:captioning}. 

\begin{table}
    \centering
    \begin{tabular}{cc}
        \toprule
        Parameter & Value \\
        \midrule
        Temperature & 1.0 \\
        top\_p & 1.0 \\
        epsilon\_cutoff & 0.01 \\
        max\_new\_tokens & 1024 \\
        \bottomrule
    \end{tabular}
    \caption{Hyperparmeters for text generation}
    \label{tab:hyperparams}
\end{table}

\begin{table}
    \centering
    \begin{tabular}{cc}
        \toprule
        Parameter & Value \\
        \midrule
        Temperature & 0.3 \\
        Version & 2024-05-13 \\
        \bottomrule
    \end{tabular}
    \caption{Hyperparmeters for GPT-4}
    \label{tab:hyperparams-gpt4}
\end{table}

\section{Reproducibility Statement}
\label{apd:reproducibility}

All datasets and models used in the experiments are publicly available. 
The code is implemented using Huggingface's Transformers library \citep{wolf-etal-2020-transformers}.
The computation of Wasserstein distance is implemented by POT: Python Optimal Transport library \cite{flamary2021pot}. 
For Section \ref{sec:mteval}, we use the codebase of \citet{vernikos2022}\footnote{\url{https://github.com/amazon-science/doc-mt-metrics}} and use MT Metrics Eval V2\footnote{\url{https://github.com/google-research/mt-metrics-eval}} for the evaluation.
Our code is available at \url{https://github.com/jinnaiyuu/mbr-optimal-transport}.

The experiments are conducted using NVIDIA A100 GPUs with 80 GB VRAM. The total amount of GPU time for the study is estimated to be 100-1000 GPU hours.

\begin{table*}
    \caption{List of datasets and models used in the experiments.}
    \label{tab:links}
    \centering
    \begin{tabularx}{\textwidth}{cX}
    \toprule
        Name & Reference \\
    \midrule
        WMT22 Metric Task & \cite{freitag-etal-2022-results} \url{https://www.statmt.org/wmt22/metrics/index.html} \\\midrule
        WMT23 Metric Task & \cite{freitag-etal-2023-results} \url{https://wmt-metrics-task.github.io/} \\\midrule
        WMT24 General Task & \cite{kocmi-etal-2024-findings} \url{https://www2.statmt.org/wmt24/translation-task.html} \\\midrule
        CNNDM & \cite{hermann2015teaching} \url{https://github.com/google-deepmind/rc-data} \\\midrule
        JADOS & \cite{nagai-etal-2024-document} \url{https://github.com/tmu-nlp/JADOS} \\\midrule
        PixelProse & \cite{singla2024pixels} \url{https://huggingface.co/datasets/tomg-group-umd/pixelprose} \\\midrule
        Llama-3.1 & \cite{dubey2024llama} \url{https://huggingface.co/meta-llama/Llama-3.1-8B-Instruct} \\\midrule
        PaliGemma-2 & \cite{steiner2024paligemma} \url{https://huggingface.co/google/paligemma2-10b-ft-docci-448} \\\midrule
        MetricX-23 & \cite{juraska-etal-2023-metricx} \url{https://huggingface.co/google/metricx-23-xxl-v2p0} \\\midrule
        MPNet & \cite{song2020mpnet} \url{https://huggingface.co/sentence-transformers/all-mpnet-base-v2} \\\midrule
        CLIP & \cite{radford2021learning} \url{https://huggingface.co/openai/clip-vit-large-patch14} \\\midrule
        COMET-22 & \cite{rei-etal-2022-comet} \url{https://huggingface.co/Unbabel/wmt22-comet-da} \\\midrule
        D-SARI & \cite{sun-etal-2021-document} \url{https://github.com/jinnaiyuu/mbr-optimal-transport} Implemented by the authors. \\\midrule
        JReadability & \cite{hasebe2015introducing} \url{https://github.com/joshdavham/jreadability} \\\midrule
        CLAIR & \cite{chan-etal-2023-clair-evaluating} \url{https://github.com/davidmchan/clair}\\
        \bottomrule
    \end{tabularx}
\end{table*}

\end{document}